\documentclass{research4cacm}
\usepackage{color}
\usepackage{overpic}
\usepackage{amsmath}

\begin{document}
%

\title{Techniques for Interpretable Machine Learning
\thanks{}
}

%
%
%
%
%
\numberofauthors{1} 
%
\author{
%
%
\alignauthor
Mengnan Du, Ninghao Liu, Xia Hu\\
       \affaddr{Department of Computer Science and Engineering, Texas A\&M University}\\
       \email{\{dumengnan,nhliu43,xiahu\}@tamu.edu}
}


\maketitle
\begin{abstract}
Interpretable machine learning tackles the important problem that humans cannot understand the behaviors of complex machine learning models and how these models arrive at a particular decision. Although many approaches have been proposed, a comprehensive understanding of the achievements and challenges is still lacking. We provide a survey covering existing techniques to increase the interpretability of machine learning models. We also discuss crucial issues that the community should consider in future work such as designing user-friendly explanations and developing comprehensive evaluation metrics to further push forward the area of interpretable machine learning. 
\end{abstract}

\section{Introduction}
Machine learning is progressing at an astounding rate, powered by complex models such as ensemble models and deep neural networks (DNNs). These models have a wide range of real-world applications, such as movie recommendations of Netflix, neural machine translation of Google, speech recognition of Amazon Alexa. 
Despite the successes, machine learning has its own limitations and drawbacks. The most significant one is the lack of transparency behind their behaviors, which leaves users with little understanding of how particular decisions are made by these models. Consider, for instance, an advanced self-driving car equipped with various machine learning algorithms doesn't brake or decelerate when confronting a stopped firetruck. 
This unexpected behavior may frustrate and confuse users, making them wonder why. Even worse, the wrong decisions could cause severe consequences if the car is driving at highway speeds and might finally crash the firetruck.
The concerns about the black-box nature of complex models have hampered their further applications in our society, especially in those critical decision-making domains like self-driving cars.

\emph{Interpretable machine learning} would be an effective tool to mitigate these problems. 
It gives machine learning models the ability to explain or to present their behaviors in understandable terms to humans~\cite{doshi2017towards}, which is named interpretability or explainability and we use them interchangeably in this paper. 
Interpretability would be an indispensable part for machine learning models in order to better serve human beings and bring benefits to society. 
For end-users, explanation will increase their trust and encourage them to adopt machine learning systems. From the perspective of machine learning system developers and researchers, the provided explanation can help them better understand the problem, the data and why a model might fail, and eventually increase the system safety. 
Thus there is a growing interest among the academic and industrial community in interpreting machine learning models and gaining insights into their working mechanisms.

Interpretable machine learning techniques can generally be grouped into two categories: intrinsic interpretability and post-hoc interpretability, depending on the time when the interpretability is obtained~\cite{molnar}. Intrinsic interpretability is achieved by constructing self-explanatory models which incorporate interpretability directly to their structures. The family of this category includes decision tree, rule-based model, linear model, attention model, etc. In contrast, the post-hoc one requires creating a second model to provide explanations for an existing model.
The main difference between these two groups lies in the trade-off between model accuracy and explanation fidelity. Inherently interpretable models could provide accurate and undistorted explanation but may sacrifice prediction performance to some extent. The post-hoc ones are limited in their approximate nature while keeping the underlying model accuracy intact.

\begin{figure}
  \centering
  \includegraphics[width=1.0\linewidth]{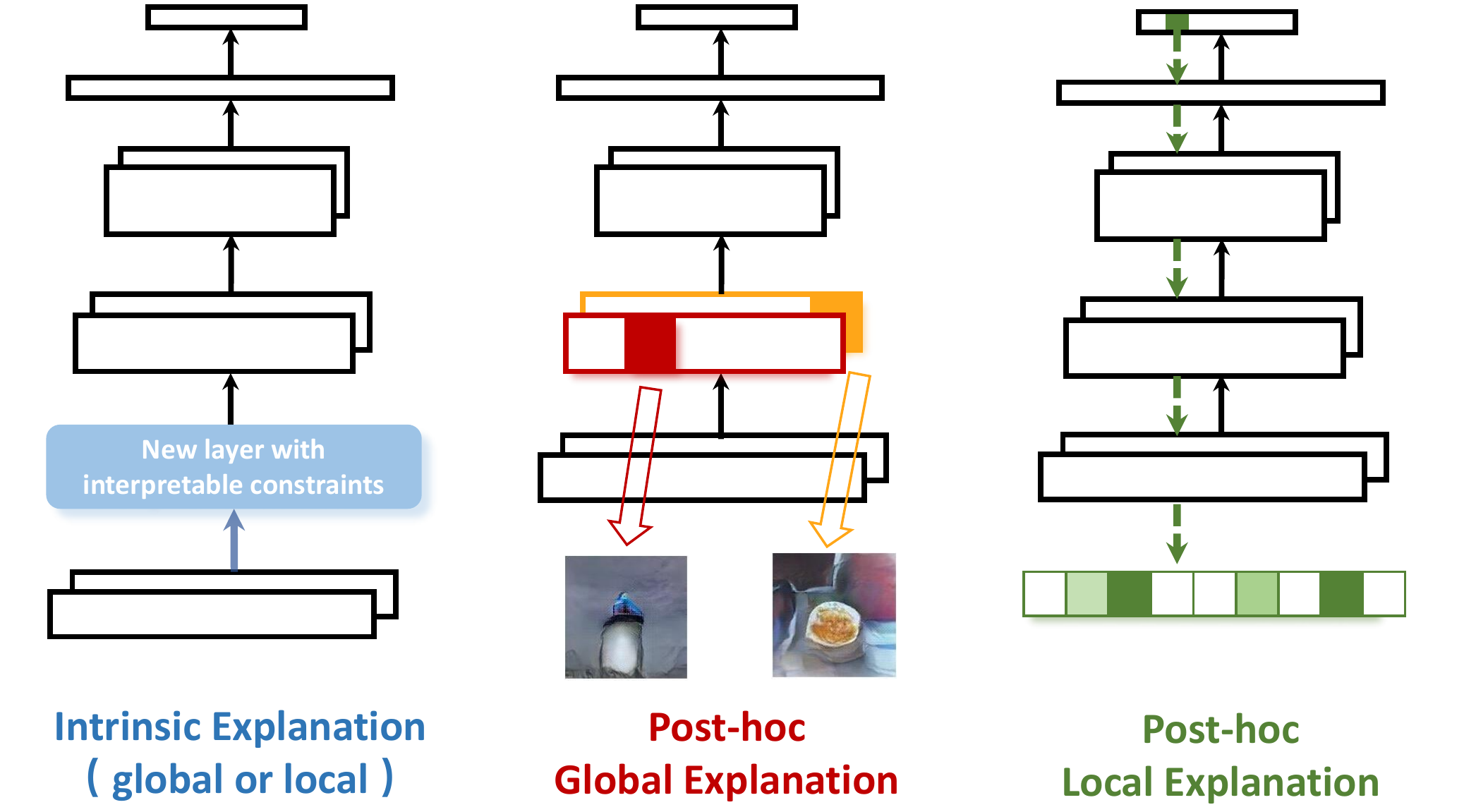}
  \caption{An illustration of three lines of interpretable machine learning techniques, taking DNN for example: Intrinsic explanation, Post-hoc global explanation of a model, and Post-hoc local explanation of a prediction.}
  \label{fig:categorization}
  \vspace{-2mm}
\end{figure}

Based on the above categorization, we further differentiate two types of interpretability: global interpretability, and local interpretability. Global interpretability means that users can understand how the model works globally by inspecting the structures and parameters of a complex  model, while local interpretability locally examines an individual prediction of a model, trying to figure out why the model makes the decision it makes. Using the DNN in Figure~\ref{fig:categorization} as an example, global interpretability is achieved by understanding the representations captured by the neurons at an intermediate layer, while local interpretability is obtained by identifying the contributions of each feature in a specific input to the prediction made by DNN. These two types bring different benefits. Global interpretability could illuminate the inner working mechanisms of machine learning models and thus can increase their transparency. Local interpretability will help uncover the causal relations between a specific input and its corresponding model prediction. Those two help users trust a model and trust a prediction, respectively.

In this article, we first summarize current progress of three lines of research for interpretable machine learning: designing inherently interpretable models (including globally and locally), post-hoc global explanation, and post-hoc local explanation. We proceed by introducing applications and challenges of current techniques. Finally, we present limitations of current explanations and propose directions towards more human-friendly explanations.

\section{Intrinsic Interpretable Model}\label{Design Interpretable Models}
Intrinsic interpretability can be achieved by designing self-explanatory models which incorporate interpretability directly into the model structures. These constructed interpretable models either are globally interpretable or could provide explanations when they make individual predictions.

\subsection{Globally Interpretable Model}

Globally interpretable models can be constructed in two ways: directly trained from data as usual but with interpretability constraints, and being extracted from a complex and opaque model.

\subsubsection{Adding Interpretability Constraints}
The interpretability of a model could be promoted by incorporating interpretability constraints. Some representative examples include enforcing sparsity terms or imposing semantic monotonicity constraints in classification models~\cite{freitas2014comprehensible}. Here sparsity means that a model is encouraged to use relatively fewer features for prediction, while monotonicity enables the features to have monotonic relations with the prediction. Similarly, decision trees are pruned by replacing subtrees with leaves to encourage long and deep trees rather than wide and more balanced trees~\cite{quinlan1987simplifying}. These constraints make a model simpler and could increase the model's comprehensibility by users.  

Besides, more semantically meaningful constraints could be added to a model to further improve interpretability. For instance, interpretable convolutional neural networks (CNN) add a regularization loss to higher convolutional layers of CNN to learn disentangled representations, resulting in filters that could detect semantically meaningful natural objects~\cite{zhang2017interpretable}. Another work combines novel neural units, called capsules, to construct a capsule network~\cite{sabour2017dynamic}.
The activation vectors of an active capsule can represent various semantic-aware concepts like position and pose of a particular object. This nice property makes capsule network more comprehensible for humans.

However, there are often trade-offs between prediction accuracy and interpretability when constraints are directly incorporated into models. The more interpretable models may result in reduced prediction accuracy comparing the less interpretable ones.

\subsubsection{Interpretable Model Extraction} 
An alternative is to apply interpretable model extraction, also referred as mimic learning~\cite{vandewiele2016genesim}, which may not have to sacrifice the model performance too much. The motivation behind mimic learning is to approximate a complex model using an easily interpretable model such as a decision tree, rule-based model, or linear model. As long as the approximation is sufficiently close, the statistical properties of the complex model will be reflected in the interpretable model. Eventually, we obtain a model with comparable prediction performance, and the behavior of which is much easier to understand. For instance, the tree ensemble model is transformed into a single decision tree~\cite{vandewiele2016genesim}. 
Moreover, a DNN is utilized to train a decision tree which mimics the input-output function captured by the neural network so that the knowledge encoded in DNN is transferred to the decision tree~\cite{bastani2017interpretability}. To avoid the overfitting of the decision tree, active learning is applied for training. These techniques convert the original model to a decision tree with better interpretability and maintain comparable predictive performance at the same time.

\subsection{Locally Interpretable Model}
Locally interpretable models are usually achieved by designing more justified model architectures that could explain why a specific decision is made. Different from the globally interpretable models that offer a certain extent of transparency about what is going on inside a model, locally interpretable models provide users understandable rationale for a specific prediction.

A representative scheme is employing attention mechanism~\cite{xu2015show,bahdanau2014neural}, which is widely utilized to explain predictions made by sequential models, e.g., Recurrent Neural Networks (RNNs). Attention mechanism is advantageous in that it gives users the ability to interpret which parts of the input are attended by the model through visualizing the attention weight matrix for individual predictions. 
Attention mechanism has been used to solve the problem of generating image caption~\cite{xu2015show}. In this case, a CNN is adopted to encode an input image to a vector, and an RNN with attention mechanisms is utilized to generate descriptions. When generating each word, the model changes its attention to reflect the relevant parts of the image. The final visualization of the attention weights could tell human what the model is looking at when generating a word. Similarly, attention mechanism has been incorporated in machine translation~\cite{bahdanau2014neural}. 
At decoding stage, the neural attention module added to neural machine translation (NMT) model assigns different weights to the hidden states of the decoder, which allows the decoder to selectively focus on different parts of the input sentence at each step of the output generation. Through visualizing the attention scores, users could understand how words in one language depend on words in another language for correct translation.

\begin{figure}
  \centering
  \includegraphics[width=0.95\linewidth]{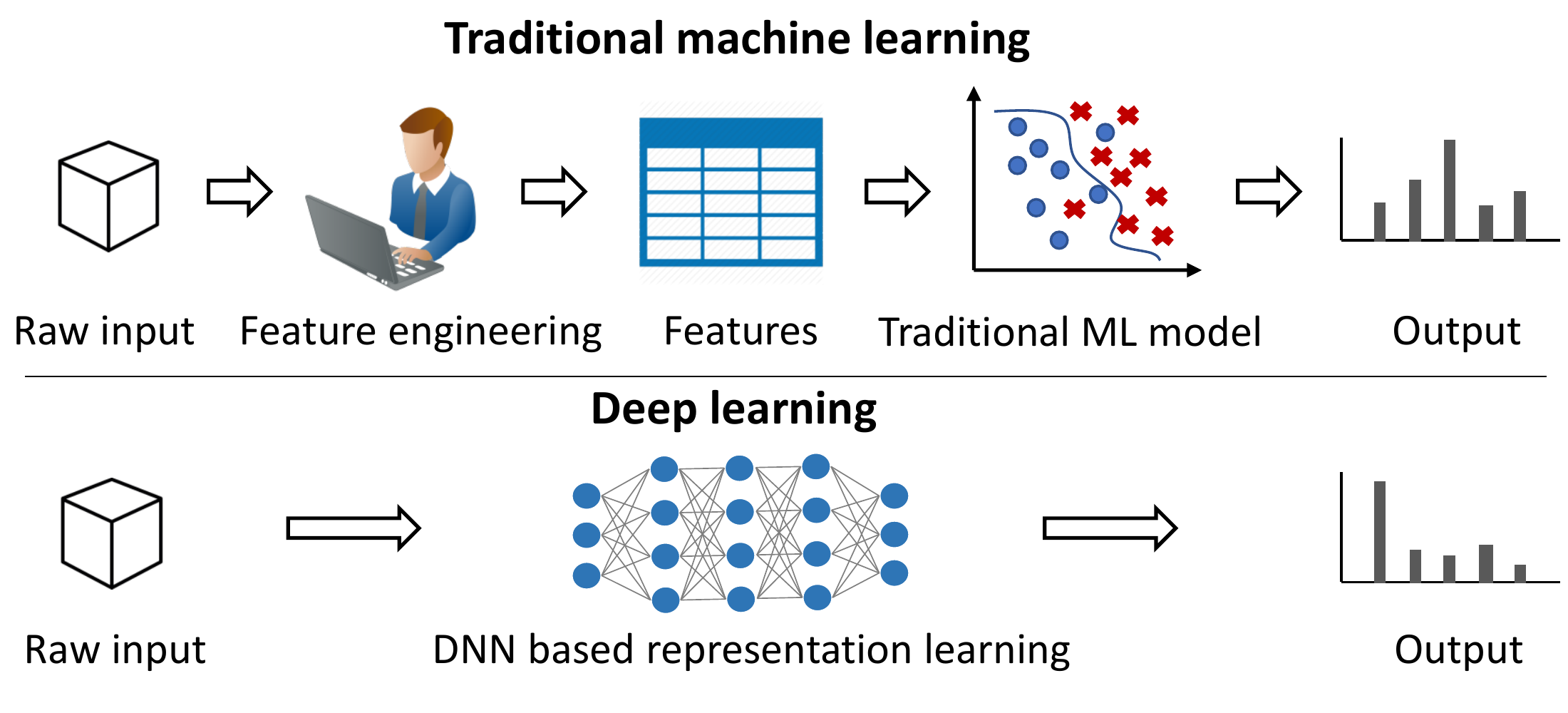}
  \caption{A traditional machine learning pipeline using feature engineering, and a deep learning pipeline using DNN based representation learning.}
  \label{fig:model-interpretation}
\end{figure}
\section{Post-hoc Global Explanation} \label{Interpretation of a Model}
Machine learning models automatically learn useful patterns from a huge amount of training data and retain the learned knowledge into model structures and parameters. Post-hoc global explanation aims to provide a global understanding about what knowledge has been acquired by these pre-trained models, and illuminate the parameters or learned representations in an intuitive manner to humans. 
We classify existing models into two categories: traditional machine learning and deep learning pipelines (see Figure~\ref{fig:model-interpretation}), since we are capable of extracting some similar explanation paradigms from each category. We introduce below how to provide explanation for these two types of pipelines.

\subsection{Traditional ML Explanation} 
Traditional machine learning pipelines mostly rely on feature engineering, which transforms raw data into features that better represent the predictive task, as shown in Figure~\ref{fig:model-interpretation}. The features are generally interpretable and the role of machine learning is to map the representation to output. 
We consider a simple yet effective explanation measure which is applicable to most of the models belonging to traditional pipeline, called \emph{feature importance}, which indicates statistical contribution of each feature to the underlying model when making decisions.

\subsubsection{Model-agnostic Explanation}
Model-agnostic feature importance is broadly applicable to various machine learning models. It treats a model as a black-box and does not inspect internal model parameters.

A representative approach is \emph{Permutation Feature Importance}~\cite{altmann2010permutation}. The key idea is that the importance of a specific feature to the overall performance of a model can be determined by calculating how the model prediction accuracy deviates after permuting the values of that feature. 
More specifically, given a pre-trained model with $n$ features and a test set, the average prediction score of the model on the test set is $p$, which is also the baseline accuracy. We shuffle the values of a feature on the test set and compute the average prediction score of the model on the modified dataset. This process is iteratively performed for each feature and eventually $n$ prediction scores are obtained for $n$ features respectively. We then rank the importance of the $n$ features according to the reductions of their score comparing to baseline accuracy $p$. There are several advantages for this approach. First, we do not need to normalize the values of the hand-crafted features. Second, it can be generalized to nearly any machine learning models with hand-crafted features as input. Third, this strategy has been proved to be robust and efficient of implementation.

\subsubsection{Model-specific Explanation}
There also exists explanation methods specifically designed for different models. 
Model-specific methods usually derive explanations by examining internal model structures and parameters. We introduce below how to provide feature importance for two families of machine learning models.

\noindent\textbf{Generalized linear models }
GLM is constituted of a series of models which are linear combination of input features and model parameters followed by feeding to some transformation function (often nonlinear)~\cite{mccullagh1989generalized}. Examples of GLM includes linear regression, logistic regression, etc. The weights of a GLM directly reflect feature importance, so users can understand how the model works by checking their weights and visualizing them. However, the weights may not be reliable when different features are not appropriately normalized and vary in their scale of measurement. Besides, the interpretability of an explanation will decrease when the feature dimensions become too large, which may be beyond the comprehension ability of humans.

\noindent\textbf{Tree-based ensemble models}
Tree-based ensemble models, such as gradient boosting machines, random forests and XGBoost~\cite{chen2016xgboost}, are typically inscrutable to humans. There are several ways to measure the contribution of each feature. The first approach is to calculate the accuracy gain when a feature is used in tree branches. The rationale behind is that without adding a new split to a branch for a feature, there may be some misclassified elements, while after adding the new branch, there are two branches and each one is more accurate. The second approach measures the feature coverage, i.e., calculating the relative quantity of observations related to a feature. The third approach is to count the number of times that a feature is used to split the data.

\subsection{DNN Representation Explanation} 
DNNs, in contrast to traditional models, not only discover the mapping from representation to output, but also learn \emph{representations} from raw data~\cite{goodfellow2016deep}, as illustrated in Figure~\ref{fig:model-interpretation}. The learned deep representations are usually not human interpretable~\cite{liu2019representation}, hence the explanation for DNNs mainly focuses on understanding the representations captured by the neurons at intermediate layers of DNNs. In the following, we introduce explanation methods for two major categories of DNN, i.e., CNN and RNN. 

\subsubsection{Explanation of CNN Representation}
There has been a growing interest to understand the inscrutable representations at different layers of CNN.
Among different strategies to understand CNN representations, the most effective and widely utilized one is through finding the preferred inputs for neurons at a specific layer. This is generally formulated in the \emph{activation maximization} (AM) framework~\cite{simonyan2013deep}, which can be formulated as:
\begin{equation}
\textbf{x}^* = \operatorname*{argmax}_{ \textbf{x} } \textbf{f}_{l} (\textbf{x}) - \mathcal{R}(\textbf{x}), 
\end{equation}
where $\textbf{f}_{l} (\textbf{x})$ is the activation value of a neuron at layer $l$ for input $\textbf{x}$, and $\mathcal{R}(\textbf{x})$ is a regularizer. Starting from random initialization, we optimize an image to maximally activate a neuron. Through iterative optimization, the derivatives of the neuron activation value with respect to the image is utilized to tweak the image. Eventually, the visualization of the generated image could tell what individual neuron is looking for in its receptive field. We can in fact do this for arbitrary neurons, ranging from neurons at the first layer all the way to the output neurons at the last layer, to understand what is encoded as representations at different layers.

While the framework is simple, getting it to work faces some challenges, among which the most significant one is the surprising artifact. The optimization process may produce unrealistic images containing noise and high-frequency patterns. Due to the large searching space for images, if without proper regularization, it is possible to produce images that satisfy the the optimization objective to activate the neuron but are still unrecognizable. To tackle this problem, the optimization should be constrained using natural image priors so as to produce synthetic images which resemble natural images. Some researchers heuristically propose hand-crafted priors, including total variation norm, $\alpha$-norm, Gaussian blur, etc. 
In addition, the optimization could be regularized using stronger natural image priors produced by a generative model, such as GAN or VAE, which maps codes in the latent space to the image spaces~\cite{nguyen2016synthesizing}. Instead of directly optimizing the image, these methods optimize the latent space codes to find an image which can activate a given neuron. Experimental results have shown that the priors produced by generative models lead to significant improvements in visualization.

The visualization results provide several interesting observations about CNN representations. First, the network learns representations at several levels of abstraction, transiting from general to task-specific from the first layer to the last layer. Take the CNN trained with the ImageNet dataset for example. Lower-layer neurons detect small and simple patterns, such as object corners and textures. Mid-layer neurons detect object parts, such as faces, legs. Higher-layer neurons respond to whole objects or even scenes. Interestingly, the visualization of the last layer neurons illustrates that CNN exhibits a remarkable property to capture global structure, local details, and contexts of an object. Second, a neuron could respond to different images that are related to a semantic concept, revealing the multifaceted nature of neurons~\cite{nguyen2016multifaceted}. For instance, a face detection neuron can fire in response to both human faces and animal faces. Note that this phenomenon is not confined to high layer neurons, all layers of neurons are multifaceted. The neurons at higher layers are more multifaceted than the ones at lower layers, indicating that higher-layer neurons become more invariant to large changes within a class of inputs, such as colors and poses. Third, CNN learns distributed code for objects~\cite{scenecnn_iclr15}. Objects can be described using part-based representations and these parts can be shared across different categories.

\subsubsection{Explanation of RNN Representation}
Following numerous efforts to interpret CNN, uncovering the abstract knowledge encoded by RNN representations (including GRUs and LSTMs) has also attracted increasing interest in recent years. Language modeling, which targets to predict the next token given its previous tokens, is usually utilized to analyze the representations learned by RNN. 
The studies indicate that RNN indeed learns useful representations ~\cite{karpathy2015visualizing,kadar2017representation,peters2018deep}. 

First, some work examines the representations of the last hidden layer of RNN and study the function of different units at that layer, by analyzing the real input tokens that maximally activate a unit. The studies demonstrate that some units of RNN representations are able to capture complex language characteristics, e.g., syntax, semantics and long-term dependencies. For instance, a study analyzes the interpretability of RNN activation patterns using character-level language modeling~\cite{karpathy2015visualizing}. This work finds that although most of the neural units are hard to find particular meanings, there indeed exist certain dimensions in RNN hidden representations that are able to focus on specific language structures such as quotation marks, brackets, and line lengths in a text. In another work, a word-level language model is utilized to analyze the linguistic features encoded by individual hidden units of RNN~\cite{kadar2017representation}. The visualizations illustrate that some units are mostly activated by certain semantic category, while some others could capture a particular syntactic class or dependency function. More interestingly, some hidden units could carry the activation values over to subsequent time steps, which explains why RNN can learn long-term dependencies and complex linguistic features. 

Second, the research finds that RNN is able to learn hierarchical representations by inspecting representations at different hidden layers~\cite{peters2018deep}. This observation indicates that RNN representations bear some resemblance to their CNN counterpart.
For instance, a bidirectional language model is constructed using a multi-layer LSTM~\cite{peters2018deep}. The analysis of representations at different layers of this model shows that the lower-layer representation captures context-independent syntactic information. In contrast, higher-layer LSTM representations encode context-dependent semantic information. The deep contextualized representations can disambiguate the meanings of words by utilizing their context, and thus could be employed to perform tasks which require context-aware understanding of words.

\section{Post-hoc Local Explanation}\label{Explanation of a prediction}

After understanding the model globally, we zoom in to the local behavior of the model and provide local explanations for individual predictions. Local explanations target to identify the contributions of each feature in the input towards a specific model prediction. As local methods usually attribute a model's decision to its input features, they are also called \emph{attribution} methods. In this section, we first introduce model-agnostic attribution methods and then discuss attribution methods specific to DNN-based predictions.

\subsection{Model-agnostic Explanation}
Model-agnostic methods allow explaining predictions of arbitrary machine learning models independent of the implementation. They provide a way to explain predictions by treating the models as black-boxes, where explanations could be generated even without access to the internal model parameters. They bring some risks at the same time, since we cannot guarantee that the explanation faithfully reflects the decision making process of a model.

\subsubsection{Local Approximation Based Explanation}

Local approximation based explanation is based on the assumption that the machine learning predictions around the neighborhood of a given input can be approximated by an interpretable white-box model. The interpretable model does not have to work well globally, but it must approximate the black-box model well in a small neighborhood near the original input. Then the contribution score for each feature can be obtained by examining the parameters of the white-box model. 

Some studies assume that the prediction around the neighborhood of an instance could be formulated as the linearly weighted combination of its input features~\cite{ribeiro2016should}. Attribution methods based on this principle first sample the feature space in the neighborhood of the instance to constitute an additional training set. A sparse linear model, such as Lasso, is then trained using the generated samples and labels. This approximation model works the same as a black-box model locally but is much easier to inspect. Finally, the prediction of the original model can be explained by examining the weights of this sparse linear model instead.

Sometimes, even the local behavior of a model may be extremely non-linear, linear explanations could lead to poor performance. Models which could characterize non-linear relationship are thus utilized as the local approximation. For instance, a local approximation based explanation framework can be constructed using if-then rules~\cite{ribeiro2018anchors}. Experiments on a series of tasks show that this framework is effective at capturing non-linear behaviors. More importantly, the produced rules are not confined merely to the instance being explained and often generalize to other instances.

\subsubsection{Perturbation Based Explanation}

This line of work follows the philosophy that the contribution of a feature can be determined by measuring how prediction score changes when the feature is altered. It tries to answer the question: which parts of the input, if they were not seen by the model, would most change its prediction? Thus, the results may be called counterfactual explanations. The perturbation is performed across features sequentially to determine their contributions, and can be implemented in two ways: omission and occlusion.
For omission, a feature is directly removed from the input, but this is impractical in practice since few models allow setting features as unknown. As for occlusion, the feature is replaced with a reference value, such as zero for word embeddings or specific gray value for image pixels. Nevertheless, occlusion raises a new concern that new evidence may be introduced and that can be used by the model as a side effect~\cite{dabkowski2017real}. For instance, if we occlude part of an image using green color and then we may provide undesirable evidence for the \emph{grass} class. Thus we should be particularly cautious when selecting reference values to avoid introducing extra pieces of evidence.

\subsection{Model-specific Explanation}   
There are also explanation approaches exclusively designed for a specific type of model. Below we introduce DNN-specific methods, which treat the networks as white-boxes and explicitly utilize the interior structure to derive explanations. We divide them into three major categories: back-propagation based methods in a top-down manner; perturbation based methods in a bottom-up manner; investigation of deep representations in intermediate layers. 

\begin{figure}
  \centering
  \includegraphics[width=0.95\linewidth]{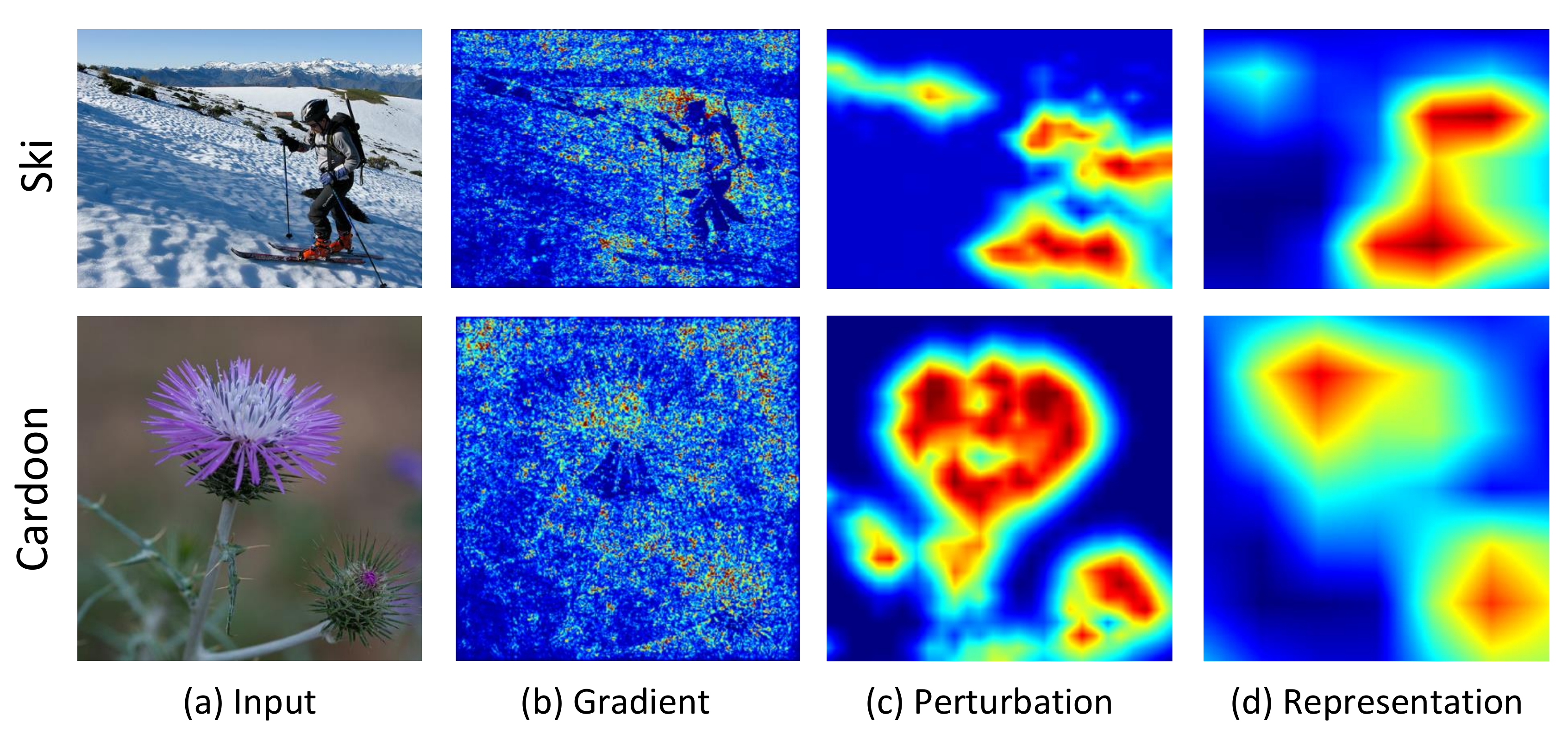}
  \caption{Local explanation heatmaps produced by (b) Back-propagation, (c) Mask perturbation, (d) Investigation of representations.}
  \label{fig:instance}
\end{figure}

\subsubsection{Back-propagation}
Back-propagation based methods calculate the gradient, or its variants, of a particular output with respect to the input using back-propagation to derive the contribution of features. In the simplest case, we can back-propagate the gradient~\cite{simonyan2013deep}.
The underlying hypothesis is that larger gradient magnitude represents a more substantial relevance of a feature to a prediction.  
Other approaches back-propagate different forms of signals to the input, such as discarding negative gradient values at the back-propagation process~\cite{springenberg2014striving}, or back-propagating the relevance of the final prediction score to the input layer~\cite{bach2015pixel}. These methods are integrated into a unified framework where all methods are reformulated as a modified gradient function~\cite{ancona2018towards}. This unification enables comprehensive comparison between different methods and facilitates effective implementation under modern deep learning libraries, such as TensorFlow and PyTorch. Back-propagation based methods are efficient in terms of implementation, as they usually need a few forward and backward calculations. On the other hand, they are limited in their heuristic nature and may generate explanations of unsatisfactory quality, which are noisy and highlight some irrelevant features, as shown in Figure~\ref{fig:instance} (b).

\subsubsection{Mask Perturbation} 
Model-agnostic perturbation mentioned in the previous section could be computationally very expensive when handling an instance with high dimensions, since they need to sequentially perturb the input. In contrast, DNN-specific perturbation could be implemented efficiently through mask perturbation and gradient descent optimization. One representative work formulates the perturbation in an optimization framework to learn a perturbation mask, which explicitly preserves the contribution values of each feature~\cite{fong2017interpretable}. Note that this framework generally needs to impose various regularizations to the mask to produce meaningful explanation rather than surprising artifacts~\cite{fong2017interpretable}. Although the optimization based framework has drastically boosted the efficiency, generating an explanation still needs hundreds of forward and backward operations. To enable more computationally efficient implementation, a DNN model can be trained to predict the attribution mask~\cite{dabkowski2017real}. Once the mask neural network model is obtained, it only requires a single forward pass to yield attribution scores for an input.

\subsubsection{Investigation of Deep Representations}
Either perturbation or back-propagation based explanations ignore the intermediate layers of the DNN that might contain rich information for interpretation. To bridge the gap, some studies explicitly utilize the \emph{deep representations} of the input to perform attribution.

Based on the observation that deep CNN representations capture the high-level content of input images as well as their spatial arrangement, a guided feature inversion framework is proposed to provide local explanations~\cite{du2018towards}. This framework inverts the representations at higher layers of CNN to a synthesized image, while simultaneously encodes the location information of the target object in a mask. Decomposition is another perspective to take advantage of deep DNN representations. For instance, through modeling the information flowing process of the hidden representation vectors in RNN models, the RNN prediction is decomposed into additive contribution of each word in the input text~\cite{du2019on}. The decomposition result could
quantify the contribution of each individual word to a RNN prediction.
These two explanation paradigms achieve promising results on a variety of DNN architectures, indicating that the intermediate information indeed contributes significantly to the attribution. Besides, deep representations serve as a strong regularizer, increasing the possibility that the explanations faithfully characterize the behaviors of DNN under normal operating conditions. Thus it reduces the risks of generating surprising artifacts and leads to more meaningful explanations.

\section{Applications}\label{Applications}
Interpretable machine learning has numerous applications. We introduce three representative ones: model validation, model debugging, and knowledge discovery.

\subsection{Model Validation}

Explanations could help to examine whether a machine learning model has employed the true evidences instead of biases which widely exist among training data. A post-hoc attribution approach, for instance, analyzes three question answering models~\cite{mudrakarta2018did}. The attribution heatmaps show that these models often ignore important part of the questions and rely on irrelevant words to make decisions.  
They further indicate that the weakness of the models is caused by the inadequacies of training data. Possible solutions to fix this problem include modifying training data or introducing inductive bias when training the model. 
More seriously, machine learning models may rely on gender and ethnic biases to make decisions~\cite{dix1992human}. Interpretability could be exploited to identify whether models have utilized these biases to ensure models don't violate ethical and legal requirements.

\subsection{Model Debugging}
Explanations also can be employed to debug and analyze the misbehavior of models when models give wrong and unexpected predictions. 
A representative example is adversarial learning~\cite{nguyen2015deep}. Recent work demonstrated that machine learning models, such as DNNs, can be guided into making erroneous predictions with high confidence, when processing accidentally or deliberately crafted inputs~\cite{nguyen2015deep,liu2018adversarial}. 
However, these inputs are quite easy to be recognized by humans. In this case, explanation facilitates humans to identify the possible model deficiencies and analyze why these models may fail. More importantly, we may further take advantage of human knowledge to figure out possible solutions to promote the performances and reasonability of models.

\subsection{Knowledge Discovery}
The derived explanations also allow humans to obtain new insights from machine learning model through comprehending their decision making process. With explanation, the area experts and the end-users could provide realistic feedbacks. Eventually, new science and new knowledge which are originally hidden in the data could be extracted. For instance, a rule-based interpretable model has been utilized to predict the mortality risk for patients with pneumonia~\cite{caruana2015intelligible}. One of the rules from the model suggests that having asthma could lower a patient's risk of dying from pneumonia. It turns out to be true since patients with asthma were given more aggressive treatments which led to better outcomes.

\section{Research challenges}\label{Research challenges}
Despite recent progresses in interpretable machine learning, there are still some urgent challenges, especially on explanation method design as well as evaluation.

\subsection{Explanation Method Design}
The first challenge is related to the method design, especially for post-hoc explanation. We argue that an explanation method should be restricted to truly reflect the model behavior under \emph{normal operation conditions}.
This criterion has two meanings. Firstly, the explanations should be faithful to the mechanism of the underlying machine learning model~\cite{du2019on}. 
Post-hoc explanation methods propose to approximate the behavior of models.
Sometimes, the approximation is not sufficiently accurate, and the explanation may fail to precisely reflect the actual operation status of the original model. For instance, an explanation method may give an explanation that makes sense to humans, while actually, the machine learning model works in an entirely different way. Second, even when explanations are of high fidelity to the underlying models, they may fail to represent the model behavior under normal conditions. Model explanation and surprising artifacts are often two sides of the same coin. The explanation process could generate examples which are out of distribution from the statistics in the training dataset, including nonsensical inputs and adversarial examples~\cite{goodfellow2014explaining}, which are beyond the capability of current machine learning models. 
Without careful design, both global and local explanations may trigger the artifacts of machine learning models, rather than produce meaningful explanations.

\subsection{Explanation Method Evaluation}
The second challenge involves the method evaluation. We introduce below the evaluation challenges for intrinsic explanation and post-hoc explanation.

The challenge for intrinsic explanation mainly lies in how to quantify the interpretability. There are broad sets of interpretable models which are designed according to distinct principles and have various forms of implementations. Take the recommender system as an example, both interpretable latent topic model and attention mechanism could provide some extent of interpretability. Nevertheless, how can we compare the interpretability between globally interpretable model and locally interpretable model? There is still no consensus on what interpretability means and how to measure the interpretability. Finale and Been propose three types of metrics: application-grounded metrics, human-grounded metrics, and functionally-grounded metrics~\cite{doshi2017towards}. These metrics are complementary to each other and bring their own pros and cons regarding the degree of validity and the cost to perform evaluations. Adopting what metrics heavily
depends on the tasks so as to make more informed evaluations.

For post-hoc explanation, comparing to evaluate its interpretability, it is equally important to assess the faithfulness of explanation to the original model, which is often omitted by existing literature. 
As mentioned before, generated explanations for a machine learning model are not always reasonable to humans.
It is extremely hard to tell whether the unexpected explanation is caused by misbehavior of the model or limitation of the explanation method. Therefore, better metrics to measure the faithfulness of explanations are needed, in order to complement existing evaluation metrics. The degree of faithfulness can determine how confident we can trust a explanation. Nevertheless, the design of appropriate faithfulness metric remains an open problem and deserves further investigation. 
\section{Discussion} 
We briefly introduce limitations of explanation methods that we have surveyed and then present explanation formats that might be more understandable and friendly to users.

\subsection{Limitations of Current Explanations}
A major limitation of existing work on interpretable machine learning is that the explanations are designed based on the intuition of researchers rather than focusing on the demands of end-users. Current local explanations are usually given in the format of feature importance vectors, which are a complete causal attribution and a low-level explanation~\cite{molnar}. This format would be satisfactory if the explanation audiences are developers and researchers, since they can utilize the statistic analysis of the feature importance distribution to debug the models. Nevertheless, this format is less friendly if the explanation receivers are lay-users of machine learning. It describes the full decision logic of a model, which contains huge amount of redundant information and will be overwhelming to users. The presentation formats could be further enhanced to better promote user satisfaction.

\subsection{Towards Human-friendly Explanations}
Based on findings in social sciences and human behavioural studies~\cite{miller2018explanation}, we provide some directions towards user-oriented explanations, which might be more satisfying to humans as a means of communication.

\vspace{2mm}
\noindent\textbf{Contrastive Explanations}
They are also referred as differential explanations~\cite{miller2018explanation}. They do not tell why a specific prediction was made, but rather explain why this prediction was made instead of another, 
so as to answer questions like \emph{``Why $Q$ rather than $R$?''}. Here $Q$ is the fact which requires explanation, and $R$ is the comparing case, which could be a real one or virtual one. Consider, for instance, a user is declined mortgage. The user may compare with another real case and raise question ``why didn't I get a mortgage when my neighborhood did?''. On the other hand, the user  
may ask ``Why was my mortgage rejected?''. Here is an implicit contrast case, and actually the user is
requesting explanation for a virtual case ``How to get my mortgage loan approved?''. Since it is compared to an event which has not happened, thus the desirable explanation here can also be called counterfactual explanation~\cite{wachter2017counterfactual}.

\begin{figure}
  \centering
  \includegraphics[width=0.95\linewidth]{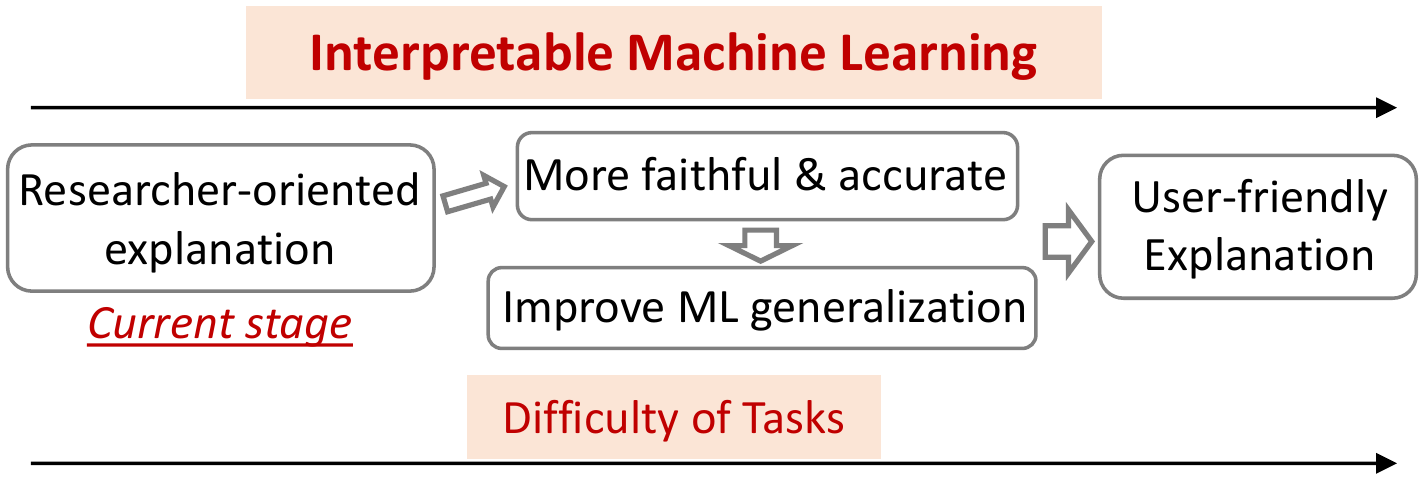}
  \caption{Progress of interpretable ML. Current stage is researcher-oriented explanations. We can make it more faithful and accurate, which can be further utilized to promote model generalization ability, and then develop user-friendly explanations.}
  \label{fig:future-of-XAI}
\end{figure}

To provide contrastive explanations for a model prediction, similar strategy could be used for both above-mentioned comparisons. We first produce feature importance attribution for two instances: not-accepted case for the user, has-accepted case of a neighbor (or would-be-accepted case of the user), and then compare the two attribution vectors. Note that we could resort to adversarial perturbation to find the would-be-accepted case. Besides, it is recommended to provide a diverse set of reasons, i.e., to find multiple contrast cases, to make the explanation more informative. Ultimately, we generate explanations of the form ``Your mortgage is rejected because your income is lower than your neighbor's, your credit history is not as strong as your neighbor's, etc'' or ``Your mortgage would be accepted if your income is raised from $x$ to $y$''. 

\vspace{2mm}
\noindent\textbf{Selective Explanations }
Usually, users do not expect an explanation can cover the complete cause of a decision. Instead, they wish the explanation could convey the most important information that contributes to the decision~\cite{miller2018explanation}. A sparse explanation, which includes a minimal set of features that help justify the prediction is preferred, although incompletely. Still use the mortgage case for example. One good explanation could be presenting users the top 2 reasons contributing to the decision, such as poor credit history, low income to debt ratio. 

\vspace{2mm}
\noindent\textbf{Credible Explanations}
Good explanation might be consistent with prior knowledge of general users~\cite{molnar}. Suppose the generated top reasons for the mortgage case include marital status is single and education status is high school graduate, then it would be less trustable than an explanation outputting poor credit history and low income to debt ratio, since the latter two are more reasonable causes leading to rejection. Low credibility could be caused by the poor fidelity of explanation to the original model. On the other hand, the explanations maybe faithful, however, the machine learning model does not adopt correct evidences to make decisions. 

\vspace{2mm}
\noindent\textbf{Conversational Explanations}
Explanations might be delivered as a conversation between the explainer and explanation receivers~\cite{miller2018explanation}. It means that we need to consider the social context, i.e., to whom an explanation is provided~\cite{tomsett2018interpretable}, in order to determine the content and formats of explanations. For instance, a preferred format is verbal explanation if it is explaining to lay-users.  

Note that there are many other paths to user-friendly explanations. We refer interested readers to the survey by Miller~\cite{miller2018explanation} for a comprehensive list of directions.
All the aforementioned directions 
serve an identical purpose that explanation should tell users why a decision was reached in a concise and friendly manner. More importantly, the explanation could inform users what could be possibly changed to receive a desired decision next time. Granted, there is still a long way to go to render explanations promote user's satisfaction. In future, 
researchers from different disciplines, including machine learning, human-computer interaction, and social science, are encouraged to closely cooperate to design really user-oriented and human-friendly explanations.

\section{Conclusions}
Interpretable machine learning is an open and active field of research, with numerous interpretation approaches continuously emerging every year. We present a clear categorization and comprehensive overview of existing techniques for  interpretable machine learning, aiming to help the community to better understand the capabilities and weaknesses of different interpretation approaches. Although techniques for interpretable machine learning are advancing quickly, some key challenges remain unsolved, and future solutions are needed to further promote the progress of this field.  


{\small\bibliographystyle{abbrv} 
\vspace{3mm}
\bibliography{bibs4cacm}}  
\balancecolumns
\end{document}